\documentclass[9pt, spconf]{article}
\usepackage{spconf}
\usepackage{graphicx}
\usepackage{caption} % DO NOT CHANGE THIS AND DO NOT ADD ANY OPTIONS TO IT
\usepackage{subcaption}
\usepackage{bm, amsmath, amssymb, amsthm}
\usepackage{booktabs}       % professional-quality tables

\def\bp{ \textbf{p} }

\def\b1{ {\bm{1}} }

\def\balpha{ \bm{\alpha} }

\def\btheta{ \bm{\theta} }

\def\nn{{ \parallel   }}

\def\EE{{ \mathbb{E}  }}

\def\diag{{ \text{diag}   }}

\def\be{{ \mathbf{e}  }}
\def\bx{{ \mathbf{x}  }}
\def\by{{ \mathbf{y}  }}

\newcommand{\defequal}{ \stackrel{\rm def}{=}  }

\title{Failure Prediction by Confidence Estimation of Uncertainty-Aware Dirichlet Networks}
\name{Theodoros Tsiligkaridis \thanks{Research was sponsored by the United States Air Force Research Laboratory and was accomplished under Cooperative Agreement Number FA8750-19-2-1000. The views and conclusions contained in this document are those of the authors and should not be interpreted as representing the official policies, either expressed or implied, of the United States Air Force or the U.S. Government. The U.S. Government is authorized to reproduce and distribute reprints for Government purposes notwithstanding any copyright notation herein.}}
\address{MIT Lincoln Laboratory \\ 244 Wood St, Lexington MA 02421 \\
    ttsili@mit.edu}
\begin{document}
%\ninept
%
\maketitle
\begin{abstract}
Reliably assessing model confidence in deep learning and predicting errors likely to be made are key elements in providing safety for model deployment, in particular for applications with dire consequences. In this paper, it is first shown that uncertainty-aware deep Dirichlet neural networks provide an improved separation between the confidence of correct and incorrect predictions in the true class probability (TCP) metric. Second, as the true class is unknown at test time, a new criterion is proposed for learning the true class probability by matching prediction confidence scores while taking imbalance and TCP constraints into account for correct predictions and failures. Experimental results show our method improves upon the maximum class probability (MCP) baseline and predicted TCP for standard networks on several image classification tasks with various network architectures.
\end{abstract}
\begin{keywords}
uncertainty, confidence, dirichlet, neural network, failure prediction
\end{keywords}

%\section{RELATION TO PRIOR WORK} \label{sec:prior}
%
%The text of the paper should contain discussions on how the paper's
%contributions are related to prior work in the field. It is important
%to put new work in  context, to give credit to foundational work, and
%to provide details associated with the previous work that have appeared
%in the literature. This discussion may be a separate, numbered section
%or it may appear elsewhere in the body of the manuscript, but it must
%be present.
%
%You should differentiate what is new and how your work expands on
%or takes a different path from the prior studies. An example might
%read something to the effect: "The work presented here has focused
%on the formulation of the ABC algorithm, which takes advantage of
%non-uniform time-frequency domain analysis of data. The work by
%Smith and Cohen \cite{Lamp86} considers only fixed time-domain analysis and
%the work by Jones et al \cite{C2} takes a different approach based on
%fixed frequency partitioning. While the present study is related
%to recent approaches in time-frequency analysis [3-5], it capitalizes
%on a new feature space, which was not considered in these earlier
%studies."

\section{Introduction} \label{intro}
Deep neural networks have achieved state-of-the-art performance  \cite{LeCun:2015} in image classification \cite{Krizhevsky:NIPS:2012, He:2015}, object detection \cite{Ren:2015, Liu:2016, Ghiasi:2019}, speech recognition \cite{Xiong:2017, Hinton:2012}, and other domains including bioinformatics \cite{Alipanahi:2015}, drug discovery \cite{Chen:2018} and medicine \cite{Wang:2016}. While deep learning algorithms and architectures are progressing, safety is an important concern when such systems are deployed in the real world \cite{Amodei:2016, Geirhos:NIPS:2018, Kelly:2019}. It is known that standard neural networks (NNs) provide poorly calibrated prediction scores \cite{Guo:ICML:2017} and overly confident predictive distributions \cite{Louizos:ICML:2017} that render them unsuitable for decision making. Understanding when a neural network model is likely to make errors is important for safe deployment in real-world conditions and high-risk applications such as healthcare, autonomous driving and cybersecurity.

We study the problem of failure prediction in deep neural networks. The widely known baseline for estimating confidence known as \textit{maximum class probability} (MCP) is inherently flawed since it assigns high confidence scores even for failure cases which deteriorates the ranking for failure prediction \cite{Nguyen:2015}. Specifically we aim to provide confidence estimates for measuring the trust in the model predictions that align well with separating correct predictions from failure cases by using a natural confidence metric given by \textit{true class probability} (TCP) \cite{Corbiere:2019}. Confidence estimates may be used for ranking input data examples and to avoid high-risk errors in uncertain scenarios by triggering human involvement or a more accurate, but possibly expensive, classifier.

In this paper, Information Aware Dirichlet (IAD) networks \cite{Tsiligkaridis:arxiv:2020} are adopted that deliver more accurate predictive uncertainty than other state-of-the-art methods by learning distributions on class probability assignments and couple them to learn the TCP confidence for the task of  failure prediction. Under a Dirichlet prior on the prediction probabilities, IAD networks are trained to minimize the approximate expected max-norm of the prediction error and align the Dirichlet concentration parameters to a direction that minimizes information captured by incorrect outcomes. Experimental results show that the TCP error score distribution has more mass concentrated towards lower scores than standard networks, which our learning criterion captures by including constraints.

The contributions of this paper are as follows. First, it is shown that IAD networks improve the separation of correct predictions and misclassifications in comparison to conventional networks. Second, a transfer-learning method is proposed for estimating the true class probability based on a constrained confidence network. Third, empirical results are provided on image classification tasks that show our method improves upon the state-of-the-art in failure case prediction.

\subsection{Related Work}
Prior work on uncertainty estimation in neural networks include Bayesian neural networks (BNN) trained with  variational inference \cite{Blundell:ICML:2015, Kingma:NIPS:2015, Molchanov:ICML:2017, Gal:ICML:2016, Li:ICML:2017}, Laplace approximation \cite{MacKay:1992, Ritter:2018}, expectation propagation \cite{Hernandez:2015, Sun:2017}, and Hamiltonian Monte Carlo \cite{Chen:2014}. Deep Dirichlet networks have been shown to outperform BNNs in uncertainty quantification for out-of-distribution and adversarial queries \cite{Sensoy:NIPS:2018, Malinin:2019, Tsiligkaridis:arxiv:2020}. These works used entropy-based metrics which measure dispersion but do not directly address discrimination of failures from correct predictions.

% specific to failure prediction
Baselines have been established for detection of misclassifications based on MCP of softmax scores  \cite{Hendrycks:2017}. Since MCP outputs high scores by default, there is significant room for improvement. A confidence metric, Trust Score, has been proposed that measures the agreement between the classifier and a modified nearest-neighbor classifier on the test example \cite{Jiang:2018}. However, this approach is not scalable due to nearest-neighbor computations and local distances in high dimensions are less meaningful \cite{Beyer:1999}. %Recent guarantees of the TCP have been established that make it attractive for failure prediction and
A confidence network has been proposed to predict TCP that empirically improved upon Trust Score and MCP \cite{Corbiere:2019}. However, results were based on networks trained with cross-entropy which are known to exhibit overconfident predictions \cite{Louizos:ICML:2017}, and treated correct predictions and failures the same, did not include TCP constraints. These drawbacks bias predicted confidence scores towards correct examples when there are much fewer errors in the training set and degrade generalization of the confidence network. In this work, constrained confidence networks are proposed that improve upon these issues by working in conjunction with information-aware Dirichlet networks.

% Recently, in \cite{Sensoy:NIPS:2018, Malinin:2019} the Dirichlet distribution was used to model distributions of class compositions and its parameters were learned by training deterministic neural networks. This approach yields closed-form predictive distributions and outperforms BNNs in uncertainty quantification for out-of-distribution and adversarial queries. However, uncertainty estimation performance for within-distribution queries was not studied and out-of-distribution and adversarial query uncertainty can be improved. The authors in \cite{Sensoy:NIPS:2018} provide a limited analysis of their loss, and \cite{Malinin:2019} lacks analysis that relates the Dirichlet concentration parameters with their loss and further proposes to use OOD data for learning what is anomalous biasing the predictive uncertainty of the models.

%In contrast, we provide a more thorough analysis of our loss function that yields insights into how neural networks shape Dirichlet distributions on the simplex. Furthermore, our method assigns higher uncertainties to errors while maintaining high confidence for correct predictions, and improves upon uncertainty quantification for adversarial data.

\section{Deep Dirichlet Networks}

\subsection{Preliminaries}
Given dataset $\mathcal{D}=\{(\bx_i,\by_i)\}_{i=1}^N$ with corresponding correct classes $c_i$, class probability vectors $\bp_i$ for sample $i$ are modeled as random vectors drawn from a Dirichlet prior distribution conditioned on the input $\bx_i$ and weights $\btheta$, i.e., $\bp_i\sim f(\cdot|\bx_i,\btheta)$. The predictive uncertainty of a classification model may be expressed as $P(y=j|\bx^*,\mathcal{D}) = \int P(y=j|\bx^*,\btheta) p(\btheta|\mathcal{D}) d\btheta = \int \int P(y=j|\bp) f(\bp|\bx^*,\btheta) d\bp \cdot p(\btheta|\mathcal{D}) d\btheta$.
%\begin{align*}
	%P(y=j|\bx^*,\mathcal{D}) &= \int P(y=j|\bx^*,\btheta) p(\btheta|\mathcal{D}) d\btheta \\
		%&= \int \int P(y=j|\bp) f(\bp|\bx^*,\btheta) d\bp \cdot p(\btheta|\mathcal{D}) d\btheta %\\
		%%&\quad = \int P(y=j|\bp) f(\bp|\bx^*,\mathcal{D}) d\bp
%\end{align*}
Distribution uncertainty $f(\bp|\bx^*,\btheta)$ is incorporated to learn priors that control the information spread on the probability simplex during the training process that in turn improve predictive uncertainty. As approximate variational inference methods, e.g, \cite{Blundell:ICML:2015, Gal:ICML:2016}, or ensemble methods may be used to estimate the intractable posterior density $p(\btheta|\mathcal{D})$, for simplicity the assumption is made that a point-estimate of the weight parameters is sufficient given a large training set and proper regularization control, thus $f(\bp|\bx^*,\mathcal{D}) = \int f(\bp|\bx^*,\btheta)p(\btheta|\mathcal{D}) d\btheta  \approx f(\bp|\bx^*,\bar{\btheta})$ similar to \cite{Sensoy:NIPS:2018, Malinin:2019}.

%Conventional NNs for classification trained with a cross-entropy loss with a softmax output layer provide a \textit{point estimate} of the predictive class probabilities of each example and do not have a handle on the underlying uncertainty. Cross-entropy training can be probabilistically interpreted as maximum likelihood estimation, which cannot infer predictive distribution variance. The softmax layer also tends to inflate the predicted class likelihood due to the exponentiation involved and this type of training tends to produce overconfident wrong predictions.

%\subsection{Deep Dirichlet Networks} \label{sec:Dirichlet}
Deep Dirichlet networks use the Dirichlet distribution \cite{Mauldon:1959, Mosimann:1962} as a prior over class probability vectors for classification tasks. Given the probability simplex as $\mathcal{S} = \left\{(p_1,\dots,p_K): p_k \geq 0, \sum_k p_k=1\right\}$, the Dirichlet distribution is $f(\bp;\balpha) = \frac{\Gamma(\alpha_0)}{\prod_{j=1}^K \Gamma(\alpha_j)} \prod_{j=1}^K p_j^{\alpha_j-1}, \bp \in \mathcal{S}$
%\begin{equation*} %\label{eq:Dirichlet}
	%f(\bp;\balpha) = \frac{\Gamma(\alpha_0)}{\prod_{j=1}^K \Gamma(\alpha_j)} \prod_{j=1}^K p_j^{\alpha_j-1}, \quad \bp \in \mathcal{S}
%\end{equation*}
where $\alpha_0=\sum_k \alpha_k$ is the Dirichlet strength parameter. The concentration parameters $\balpha = (\alpha_1,\dots,\alpha_K)$ are larger than unity \footnote{This constraint avoids inversion of the Dirichlet distribution as $\alpha_k<1$ concentrates mass in the simplex corners and its boundaries.} and may be interpreted as how likely a class is relative to others. The all-ones $\balpha$ vector reduces to the uniform distribution over the simplex. A Dirichlet neural network's output layer determines the concentration parameters $\balpha = g_\alpha(\bx^*; \bar{\btheta}) + 1$, the final activation is a softplus activation layer that outputs non-negative continuous values. These parametrize the density $f(\bp|\bx^*, \bar{\btheta}) = f(\bp; \balpha)$ and the predictive distribution is	%$P(y=j|\bx^*;\bar{\btheta}) = \EE_{f(\bp|\bx^*;\bar{\btheta})}[P(y=j|\bp)] = \frac{\alpha_j}{\alpha_0}$
%\begin{equation*}
$P(y=k|\bx^*;\bar{\btheta}) = \EE_{\bp \sim f(\cdot|\bx^*;\bar{\btheta})}[P(y=k|\bp)] = \alpha_k/\alpha_0$
%\end{equation*}
with predicted class $\hat{y} = \arg \max_k P(y=k|\bx^*;\bar{\btheta})$.

\subsection{Training Framework}
In contrast to cross-entropy (CE) training that only seeks to maximize the correct class likelihood, information-aware deep Dirichlet networks \cite{Tsiligkaridis:arxiv:2020} learn the Dirichlet prior concentration parameters by minimizing an approximate Bayes risk of the prediction error in $L_\infty$ space given by\footnote{The larger $p$, the tighter the $L_p$ norm approximated the max-norm.}
\begin{align*}
    \mathcal{F}_i &= \left( \EE_{\bp_i \sim f(\cdot;\balpha_i)}\left[\sum_k (y_{ik}-p_{ik})^p\right] \right)^{1/p} \\
    &\geq \EE_{\bp_i \sim f(\cdot;\balpha_i)}[\nn \by_i-\bp_i \nn_p] \geq \EE_{\bp_i \sim f(\cdot;\balpha_i)}[\nn \by_i-\bp_i \nn_\infty]
\end{align*}
in addition to minimizing the information captured associated with incorrect outcomes, $\mathcal{R}_i$:
\begin{align}
	\mathcal{L}(\btheta) &= \frac{1}{N}\sum_{i=1}^N \mathcal{F}_i + \lambda \mathcal{R}_i   \label{eq:IRD_loss} \\
	\mathcal{F}_i &\defequal \frac{1}{\mu(\alpha_{i0})^{1/p}} \Big( \mu(\sum_{k\neq c_i} \alpha_{ik}) + \sum_{k\neq c_i} \mu(\alpha_{ik}) \Big)^{1/p} \nonumber \\
	\mathcal{R}_i &\defequal \frac{1}{2} \sum_{k\neq c_i} (\alpha_{ik}-1)^2 [J(\tilde{\balpha}_i)]_{kk} \nonumber
\end{align}
where $\lambda>0$ is a regularization parameter \footnote{A gradual annealing $\lambda_t = \lambda \min\{\frac{t-T_0}{T}, 1\} I_{\{t>T_0\}}$ allows the network to learn good features for classification before introducing the regularizer.}, $\mu(\alpha)\defequal \Gamma(\alpha+p)/\Gamma(\alpha)$, $\psi(\cdot)$ is the digamma function and $J(\tilde{\balpha}) = -\EE_{\bp\sim f(\cdot;\tilde{\balpha})}[\nabla^2 \log f(\bp; \tilde{\balpha})] = \diag(\{\psi^{(1)}(\tilde{\alpha}_{ik})\}_k) - \psi^{(1)}(\tilde{\alpha}_{i0}) \cdot 1_{K\times K}$
%\begin{align*}
    %J(\tilde{\balpha}) &= -\EE_{\bp\sim f(\cdot;\tilde{\balpha})}[\nabla^2 \log f(\bp; \tilde{\balpha})] \\
    %&= \diag(\{\psi^{(1)}(\tilde{\alpha}_{ik})\}_k) - \psi^{(1)}(\tilde{\alpha}_{i0}) 1_{K\times K} 
%\end{align*}
is the Fisher information matrix of the Dirichlet density associated with the modified concentration vector $\tilde{\balpha}_i = (1-\by_i) \odot \balpha_i + \by_i$ that ties the correct class concentration parameter $\alpha_{c_i}$ to unity. It is a local approximation to the R\'enyi divergence between densities $f(\bp;\tilde{\balpha})$ and $f(\bp;\b1)$.

Several analytical properties of this loss have been established; (a) $\mathcal{F}_i$ is decreasing in $\alpha_c$, (b) $\mathcal{F}_i$ is increasing in $\alpha_k,k\neq c$ as $\alpha_k$ grows; that imply by minimizing this the model (a) encourages the class probability vectors to concentrate towards the correct class and (b) avoids assigning high concentration parameters to incorrect classes since observations that are assigned incorrect outcomes cannot be explained \cite{Tsiligkaridis:arxiv:2020}. While the classification loss, $\mathcal{F}_i$, achieves high accuracy by learning interesting patterns, the network may learn that certain patterns contribute to large information flow towards misleading classes which affects predictive uncertainty. The information regularization $\mathcal{R}_i$ is increasing in $\alpha_k, k\neq c$, implying that it further minimizes information associated with misleading outcomes \cite{Tsiligkaridis:arxiv:2020}.

The IAD loss improves upon the mean-square-error loss $\EE_{\bp_i\sim f(\cdot;\balpha_i)}[\nn \by_i-\bp_i \nn_2^2]$ proposed in \cite{Sensoy:NIPS:2018} as the $L_\infty$ norm minimizes the cost of the highest prediction error among the classes, while the $L_2$ norm minimizes the sum-of-squares easily affected by outlier scores, $\nn \be_i \nn_\infty \leq \nn \be_i \nn_p \leq \nn \be_i \nn_2$ for $p>2$, and as a result when errors occur the uncertainty is expected to be higher as the effect of favoring one class more than others is mitigated. It also improves upon the  KL-loss from \cite{Malinin:2019} $D_{KL}( f(\cdot;\alpha_i) \parallel f(\cdot;(\beta+1)\by_i + (1-\by_i)))$ for some arbitrary $\beta$ as it tries to fit the best Dirichlet prior to each training example (since one cannot expect all examples to yield highly-concentrated Dirichlet prior distributions); perhaps more importantly it does not rely on access to out-of-distribution data at training time.

%\begin{theorem} \label{thm:IRD}
	%The loss function $L_i(\theta)$ is (a) decreasing in $\alpha_c$ and (b) increasing in $\alpha_k, k\neq c$.
%\end{theorem}
%\begin{proof}
%For the first part, we observe:
%\begin{equation*}
	%\frac{\partial\mathcal{F}_i}{\partial \alpha_{i,c_i}}=\psi^{(1)}(\alpha_{i,0})-\psi^{(1)}(\alpha_{i,c_i}) < 0
%\end{equation*}
%which follows from Lemma 2 in \cite{Tsiligkaridis:arxiv:2020}. The second term, $\mathcal{R}_i$, is independent of $\alpha_{i,c_i}$. For the second part, we observe for $k\neq c_i$:
%\begin{equation*}
	%\frac{\partial\mathcal{F}_i}{\partial \alpha_{i,k}}=\psi^{(1)}(\alpha_{i,0}) > 0
%\end{equation*}
%which follows from $\psi^{(1)}(z+1)=\int_0^1 \frac{t^z}{1-t} \log(1/t) dt >0$, implying that the $\mathcal{F}_i$ is increasing in $\alpha_k$. Theorem 3 in \cite{Tsiligkaridis:arxiv:2020} shows $\mathcal{R}_i$ is increasing in $\alpha_k$, thus the sum $\mathcal{F}_i+\lambda \mathcal{R}_i$ is also increasing.
%\end{proof}
%Theorem \ref{thm:IRD} implies that imply that through minimization of our (a) encourages the class probability vectors to concentrate towards the correct class and (b) avoids assigning high concentration parameters to incorrect classes since observations that are assigned incorrect outcomes cannot be explained.

\section{Learning Model Confidence}

%\subsection{True Class Probability}
The true class probability (TCP) has been recently proposed as an improvement to  max-class probability (MCP), given by $\max_j P(y=j|\bx_i;\bar{\btheta})$ which leads to high-confidence values making it difficult to predict failure cases \cite{Corbiere:2019}. The true class probability (TCP) score $c^*(\bx_i,c_i) = \frac{\alpha_{i,c_i}}{\alpha_{i,0}}$ associated with a classifier has certain theoretical guarantees \cite{Corbiere:2019}: Given example $(x_i,y_i)$, (a) when $c^*(x_i,c_i)>1/2$, the predicted label is correct, i.e., $\hat{y}_i=y_i$, and (b) when $c^*(x_i,c_i)<1/K$ the example is misclassified $\hat{y}_i \neq y_i$.

\begin{figure}
\centering
\begin{subfigure}{.25\textwidth}
  \centering
  \includegraphics[width=1.0\textwidth]{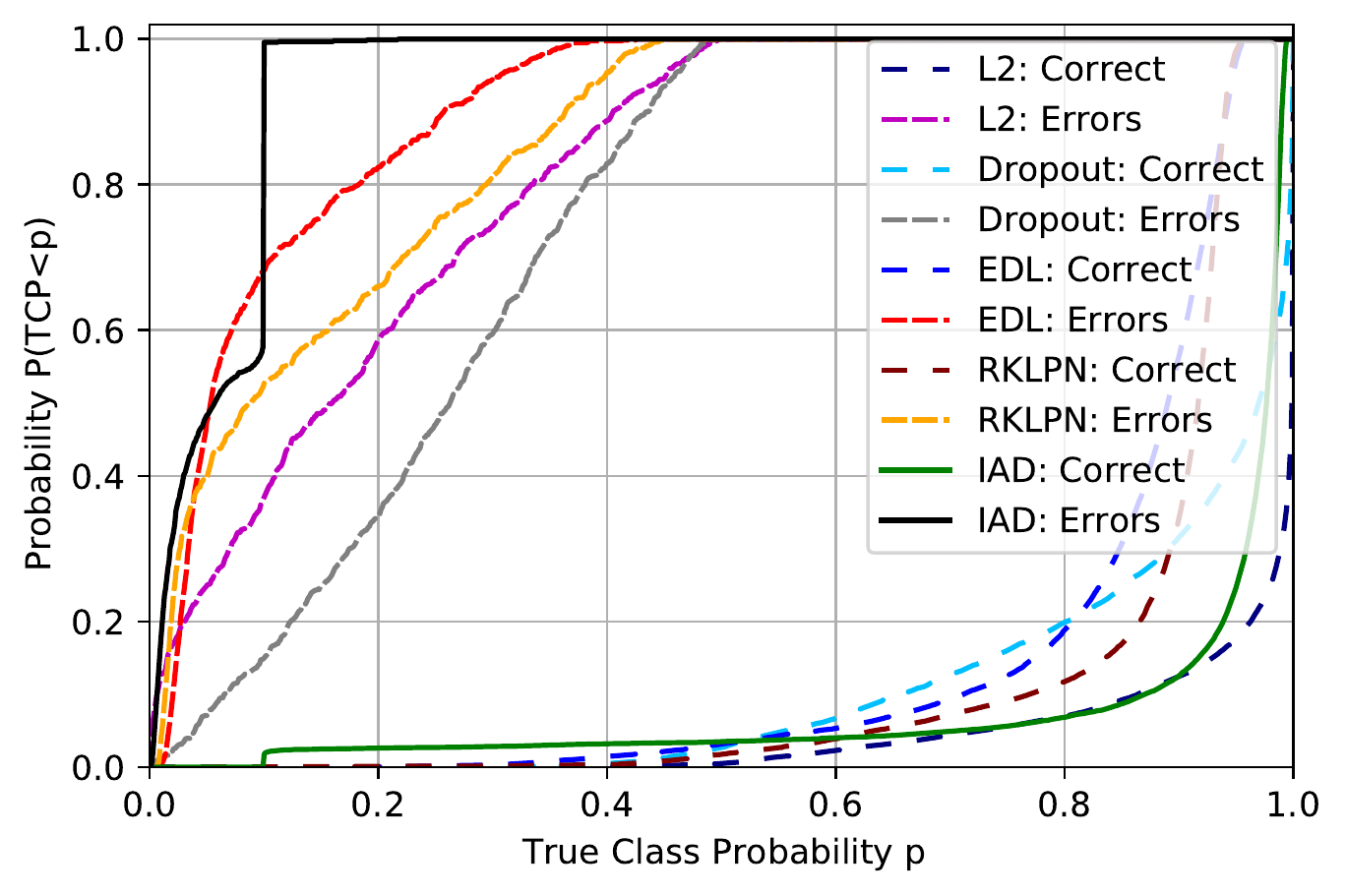}
%  \caption{A subfigure}
  \label{fig:sub1}
\end{subfigure}%
\begin{subfigure}{.25\textwidth}
  \centering
  \includegraphics[width=1.0\textwidth]{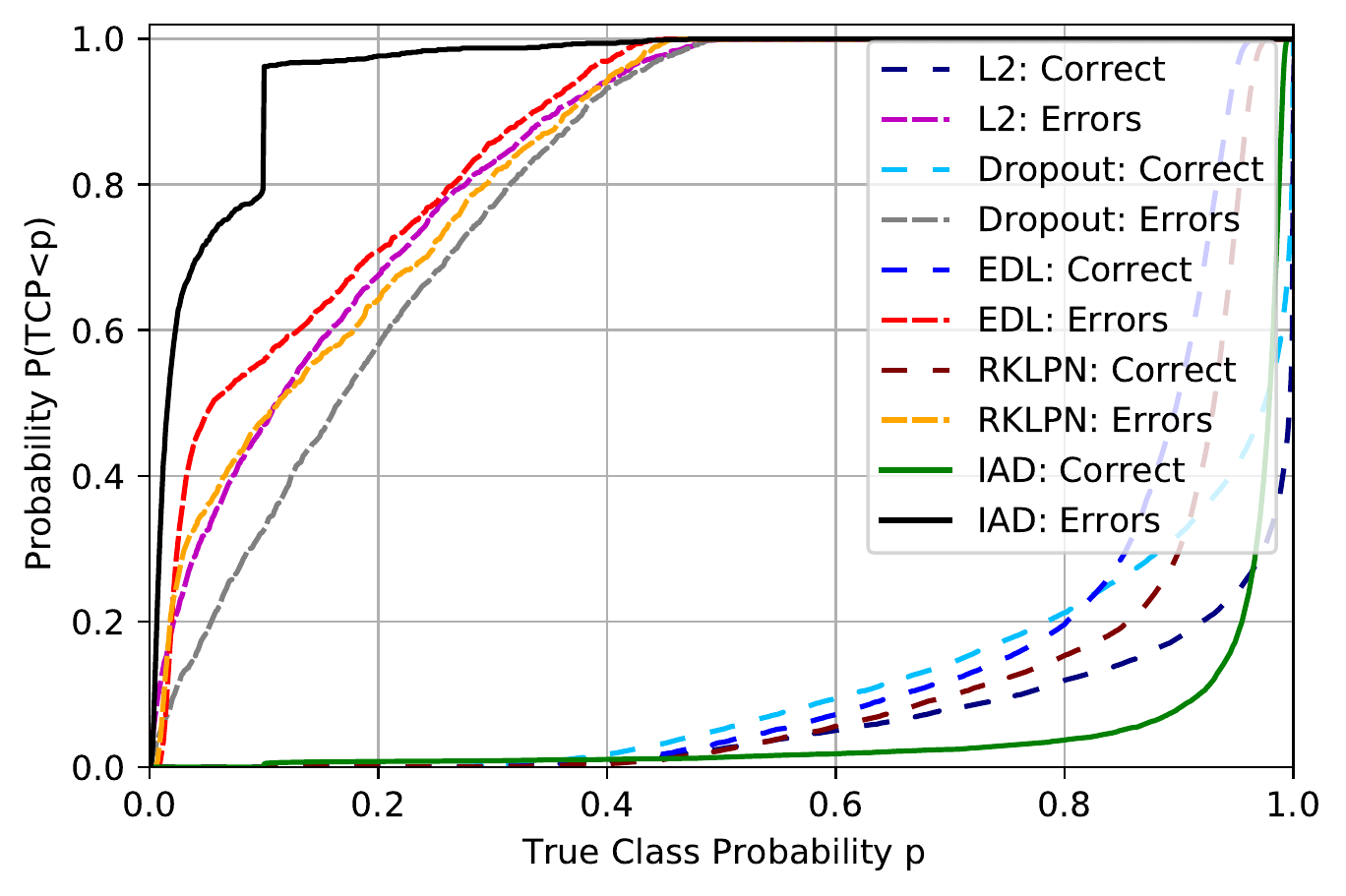}
%  \caption{A subfigure}
  \label{fig:sub2}
\end{subfigure}
\caption{Empirical cumulative density function (CDF) of TCP scores for various deep learning methods on Fashion-MNIST (left) and CIFAR-10 (right) test sets.}
\label{fig:cdf_tcp}
\end{figure}

%For deep Dirichlet networks, we use TCP $c^*(x_i,c_i)$ as a confidence metric for failure prediction.
%v^*(\bx_i,c_i) &= \Var_{\bp \sim f(\cdot|\bx_i;\bar{\btheta})}[P(y=c_i|\bp)] = \frac{\alpha_{i,c_i}(\alpha_{i,0}-\alpha_{i,c_i})}{\alpha_{i,0}^2(\alpha_{i,0}+1)}
Figure \ref{fig:cdf_tcp} shows the empirical cumulative density functions for the Fashion-MNIST and CIFAR-10 datasets of the TCP associated with correct predictions and errors for several networks: CE-trained networks with weight decay, Monte Carlo-Dropout \cite{Gal:ICML:2016}, reverse KL divergence-prior network (RKLPN) \cite{Malinin:2019}, evidential DL (EDL) \cite{Sensoy:NIPS:2018}, and information-robust Dirichlet (IAD) network \cite{Tsiligkaridis:arxiv:2020}. Unlike all other methods, the IAD method stands out by distributing most of its error TCP scores below $1/K$, which is linked to the theoretical regime of misclassifications, while maintaining high TCP scores for correct predictions. This can be attributed to the properties of $\mathcal{F}_i, \mathcal{R}_i$ that minimize information associated with incorrect outcomes. Since $\alpha_c < \max_{k\neq c} \alpha_k$ for failure cases, this preferable behavior leads to lower TCP scores.

% \begin{figure}[htbp]
% 	\centering
% 		\includegraphics[width=0.50\textwidth]{./Figs/cdf_tcp_joint.jpg}
% 	\caption{Empirical CDF of TCP scores for various deep learning methods on Fashion-MNIST (a) and CIFAR-10 (b) test sets.}
% 	\label{fig:cdf_tcp}
% \end{figure}
While the TCP criterion offers great separation between failure cases and correct predictions, the failure prediction performance hinges on how well these TCP scores can be estimated as the true class is unknown in practice. We propose a constrained confidence network with parameters $\theta_c$ that learns TCP scores subject to boundary constraints by minimizing
\begin{align}
	&\mathcal{L}_c(\btheta_c) = \frac{1}{N} \sum_{i=1}^N \Big\{ (1-s_i) \left( \text{MSE}(\bx_i;\btheta_c) + \lambda_c \phi(\bx_i,-M,t_c;\btheta_c) \right) \nonumber \\
	&\quad + \zeta s_i \left( \text{MSE}(\bx_i;\btheta_c) + \lambda_c \phi(\bx_i,M,t_e;\btheta_c) \right) \Big\}  \label{eq:Lc} 
	%\sigma(-M (\hat{c}(\bx_i,\btheta_c)-t_c)) \nonumber \\
	%\sigma(M (\hat{c}(\bx_i,\btheta_c)-t_e))
\end{align}
where $\text{MSE}(\bx_i;\btheta_c) = \left(\hat{c}(\bx_i,\btheta_c)-c^*(\bx_i,c_i)\right)^2$, $\phi(\bx_i,M,t;\btheta_c) = \sigma(M (\hat{c}(\bx_i,\btheta_c)-t))$, $\sigma(x)=1/(1+e^{-x})$ is the sigmoid function, and $M\gg 1$ controls the smoothness of the constraint functions $\phi$. Parameters $\lambda_c>0$ controls constraint enforcement and $\zeta>0$ balances between misclassified and correctly classified predictions. Here, $s_i=I_\{\hat{y}_i\neq y_i\}=1$ for misclassified examples $\bx_i$ and zero otherwise. The thresholds $t_e,t_c$ are chosen as $t_e=1/K+\delta,t_c=t_e+\epsilon$ for small margins $\delta,\epsilon>0$. These constraints lead to feasible solutions for the IAD method due to the sharp transition near $1/K$.

\begin{figure}[t]
	\centering
		\includegraphics[width=0.50\textwidth]{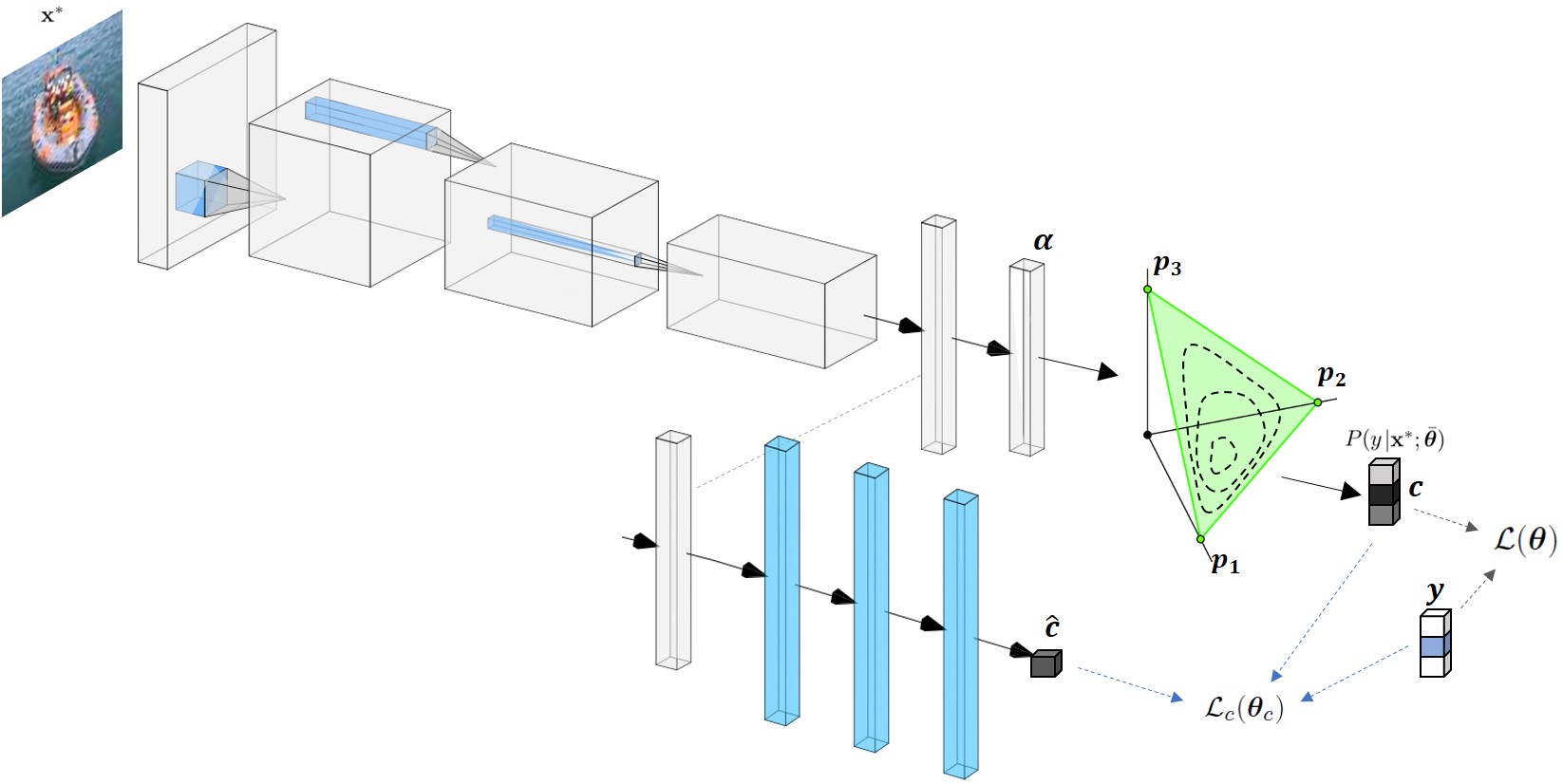}
	\caption{The constrained confidence network (blue) transfers knowledge from the primary classification network (gray) to learn confidence scores $\hat{c}(\bx,\btheta_c)$.}
	\label{fig:concept}
\end{figure}

The confidence network reuses the convolutional layers of the Dirichlet network by freezing them and adds several dense layers with a penultimate dense layer with sigmoid activation to form $\hat{c}(\bx_i,\btheta_c)$ (see Fig. \ref{fig:concept}). \footnote{Fine-tuning the full architecture is also possible but is not considered here.} Our goal is not to change the original classification model's accuracy, but to predict its failure cases. In experiments, we implemented other methods including weighted mean-square error regression, contrastive loss, binary cross entropy and its focal loss variant. However, the constrained mean-square error regression (\ref{eq:Lc}) achieved the best failure predictions performance.
%and a dense layer with softplus activation to form $\hat{v}(\bx_i,\btheta_c)$. %\footnote{Fine-tuning the full architecture is also possible but is not considered here.}

\section{Experiments}
Performance on detection of failure cases is evaluated for image classification. As the recent work \cite{Corbiere:2019} has shown that CE-TCP with a confidence network outperforms CE-MCP baseline \cite{Hendrycks:2017} and Trust Score \cite{Jiang:2018}, we focus on comparing standard CE NNs with IAD NNs with MCP or predicted TCP confidence scores.

\textit{Datasets \& Architectures.} Our evaluation spans an array of different resolution and increasing complexity datasets. These include Fashion-MNIST (grayscale $28\times 28$-$10$ classes), CIFAR-10 (color $32\times 32$-$10$ classes), CIFAR-100-coarse (color $32\times 32$-$20$ classes), Tiny-ImageNet (color $64\times 64$-$200$ classes). The primary network architectures are contained in Table \ref{table-results}, where $n\{k\}$ denotes a filter blocks of size $n$ and kernel size $k\times k$, and $n$ denotes a dense layer of size $n$. %A LeNet CNN architecture was used for Fashion-MNIST composed of $20,50$ filters of size $5\times 5$ with $500$ hidden units at the dense layer. For CIFAR-10, a VGG-type architecture was used consisting of three filter blocks with $64,128,256$ filters of kernel size $3\times 3$ respectively and $256$ hidden units at the dense layer. For CIFAR-100-coarse, a VGG-type architecture was used made up of four filter blocks consisting of $128,256,512,512$ filters of kernel size $3\times 3$ and two dense layers of size $1024$. For Tiny-ImageNet, a VGG-type architecture was used composed of five filter blocks with $128,256,512,512,512$ filters of kernel size $3\times 3$ respectively and two dense layers of size $1024$. \footnote{For datasets, except for Fashion-MNIST, data augmentation, dropout and batch-normalization was used for all methods to mitigate overfitting.}
The confidence networks for all methods were based on three dense layers with ReLU nonlinearities made up of $n_h$ hidden nodes ($n_h=600$ for Fashion-MNIST and CIFAR-10, $n_h=1000$ for CIFAR-100-coarse and Tiny-ImageNet) followed by a dense layer with a sigmoid activation that outputs the true class probability score. For all datasets we used $\lambda_c=5$ and $\zeta=1$ (except for Tiny-Imagenet $\lambda_c=0.5$). %These hyperparameters were chosen with minimal experimentation and it's possible to obtain improved performance with further tuning.

\textit{Performance metrics.} We measure the failure case detection performance using standard metrics \cite{Hendrycks:2017}: AUROC, AUPRC-Success, AUPRC-Error, FPR at $85\%$ TPR. The main metrics of interest are AUPRC-Error which measures the area under the precision-recall curve when errors are taken as the positive class, and FPR at $85\%$ TPR which is interpreted as the probability that an error is misclassified as a correct prediction when the true positive rate is $85\%$.

%All experiments are implemented in Tensorflow \cite{Abadi:2016:TSL:3026877.3026899} and the Adam \cite{Kingma:2015} optimizer was used for training. As recent prior works \cite{Sensoy:NIPS:2018, Malinin:2019} have shown Dirichlet NNs outperform BNNs on several benchmark image datasets, we mainly focus on comparing our method with these Dirichlet NNs trained with different loss functions. Comparisons are made with the following methods: (a) L2 corresponds to deterministic neural network with softmax output and weight decay, (b) Dropout is the uncertainty estimation method of \cite{Gal:ICML:2016}, (c) EDL is the evidential approach of \cite{Sensoy:NIPS:2018}, (d) RKLPN is the reverse KL divergence-based prior network method of \cite{Malinin:2019}, and (e) IRD is our proposed technique.

\begin{figure}[h]
	\centering
	    %% left lower right upper
		%\includegraphics[trim={0 0 0 1cm},clip,width=0.45\textwidth]{./Figs/cifar10_est_tcp_hist.pdf}
		\includegraphics[width=0.47\textwidth]{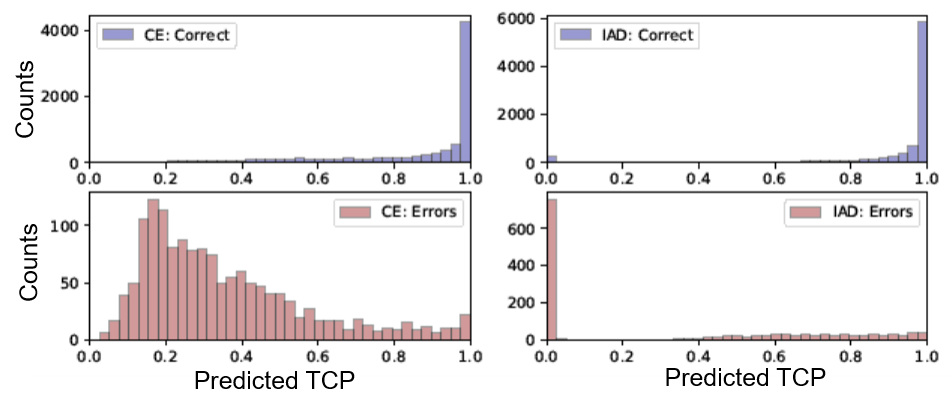}
	\caption{Histogram of predicted TCP scores for CE confidence network \cite{Corbiere:2019} (left) and our proposed IAD constrained confidence network (right) on CIFAR-10 test set. }
	\label{fig:relative_histogram_cifar10}
\end{figure}

% results
Table \ref{table-results} shows comparative results with CE NNs. We observe that IAD coupled with a TCP-constrained confidence network offers the best failure prediction performance and interestingly IAD-MCP improves considerably upon the popular CE-MCP. While AUROC and AUPRC-Success are slightly higher occasionally for CE-TCP than IAD-TCP, we remark that AUROC is sensitive to imbalance and AUPRC-Success measures detection of model successes. This may be attributed to the unconstrained TCP learning criterion of CE-TCP that comes at the cost of more dispersed clusters of scores. Figure \ref{fig:relative_histogram_cifar10} shows IAD-TCP offers significantly reduced overlap between correct predictions and failures in comparison to \cite{Corbiere:2019} and error scores are tightly clustered near zero.
\begin{table}[h]
\caption{ Comparison of failure prediction methods on various image classification datasets and network architectures. }
\label{table-results}
%\vskip 0.15in
\begin{center}
\begin{small}
\begin{sc}
\scalebox{0.9}{
\begin{tabular}{ p{60pt} p{30pt} p{30pt} p{30pt} p{40pt} }
\toprule
Method & AUROC & AUPRC-Success & AUPRC-Error & FPR at $85\%$ TPR \\
\midrule
\multicolumn{5}{l}{\emph{Fashion-MNIST}: LeNet 20\{5\}-50\{5\}-500} \\
CE-MCP    & 91.85 & 99.23 & 47.09 & 16.69 \\
CE-TCP  & 91.94 & 99.23 & 46.36 & 15.68 \\
IAD-MCP  &  92.01 & 99.05 & 61.59 & 16.65 \\
IAD-TCP* & \textbf{93.42} & \textbf{99.26} & \textbf{63.80} & \textbf{13.87} \\
\bottomrule
\multicolumn{5}{l}{\emph{CIFAR-10}: VGG 64\{3\}-128\{3\}-256\{3\}-256} \\
CE-MCP   & 89.22 & 97.97 & 51.99 & 24.61 \\
CE-TCP   & 92.92 & 98.59 & 71.55 & 13.93 \\
IAD-MCP  & 90.54 & 98.29 & 61.45 & 20.69 \\
IAD-TCP* & \textbf{93.79} & \textbf{98.91} & \textbf{74.31} & \textbf{10.81} \\
\bottomrule
\multicolumn{5}{l}{\emph{CIFAR-100-coarse}: VGG 128\{3\}-256\{3\}-512\{3\}-512\{3\}-1024-1024} \\
CE-MCP   & 86.78 & 95.72 & 62.62 & 33.33 \\
CE-TCP   & \textbf{88.71} & \textbf{96.27} & 69.36 & 27.43 \\
IAD-MCP  & 86.48 & 95.12 & 67.48 & 33.33 \\
IAD-TCP* & 88.13 & 95.63 & \textbf{73.15} & \textbf{26.38} \\
\bottomrule
\multicolumn{5}{l}{\emph{Tiny-ImageNet}: VGG 128\{3\}-256\{3\}-512\{3\}-512\{3\}-1024-1024} \\
CE-MCP   & 84.90 & 85.68 & 83.47 & 35.65 \\
CE-TCP  & \textbf{87.06} & \textbf{87.85} & 85.85 & 30.30 \\
IAD-MCP  & 82.97 & 81.55 & 83.85 & 40.29 \\
IAD-TCP* & 86.96 & 85.62 & \textbf{87.30} & \textbf{29.72} \\
\bottomrule
\end{tabular}}
\end{sc}
\end{small}
\end{center}
\vskip -0.1in
\end{table}

\section{CONCLUSION}
A new method is presented for estimating confidence of deep Dirichlet neural networks. Our method transfers knowledge from the classification network to a confidence network that learns the true class probability (TCP) by matching the TCP scores while taking imbalance and TCP constraints into account for correct predictions and errors. Empirical results show significant improvements in failure prediction for image classification tasks over state-of-the-art methods that validate the benefits of our method.

\newpage
\vfill\pagebreak

\bibliographystyle{IEEEbib}
\bibliography{refs}

\end{document}